\documentclass{article}
\usepackage{graphicx} % Required for inserting images
\usepackage[paperheight=11in,
   paperwidth=8.5in,
   top=1in,
   bottom=1in,
   left=1in,
   right=1in]{geometry}
\usepackage{amsmath}
\usepackage{amsfonts}
\usepackage{amssymb}
\usepackage{verbatim}
\usepackage{times}
\usepackage{setspace}
\usepackage{parskip}
\usepackage{caption}
\usepackage{subcaption}
\usepackage{multirow}
\usepackage{hyperref}
\usepackage{xcolor}
\usepackage{titlesec}
\titleformat{\section}
  {\fontsize{10}{12}\selectfont\bfseries} % 14pt font, 16pt line spacing, bold
  {\thesection}{1em}{}
\titleformat{\subsection}
{\fontsize{10}{12}\selectfont\bfseries} % 14pt font, 16pt line spacing, bold
{\thesubsection}{1em}{}

\usepackage[numbers,super]{natbib} % Use superscript for citations
\makeatletter
\renewcommand\@biblabel[1]{#1.} % Removes brackets from bibliography
\makeatother

\providecommand{\keywords}[1]
{
  \small	
  \textbf{\textit{Keywords---}} #1
}

\title {\fontsize{14}{10}\selectfont \textbf{Transfer Learning with Clinical Concept Embeddings from Large Language Models}}
\title{\fontsize{14}{16}\selectfont \textbf{Transfer Learning with Clinical Concept Embeddings from Large Language Models}}
\author{\fontsize{12}{14} \textbf{Yuhe Gao$^{1}$, MS, Runxue Bao$^{2}$, PhD, Yuelyu Ji$^{1}$, MS, Yiming Sun$^{1}$, BE,} \\
        \fontsize{12}{14} \textbf{Chenxi Song$^{1}$, MS, Jeffrey P Ferraro$^{3}$, PhD, Ye Ye$^{1*}$, PhD} \\
        \fontsize{12}{14} \textbf{$^{1}$University of Pittsburgh; $^{2}$GE Healthcare; $^{3}$University of Utah;} \\
        \fontsize{12}{14} \textbf{$^{*}$Corresponding Author}}
\date{}

\begin{document}

\maketitle

\begin{abstract}
\noindent
\textit{Knowledge sharing is crucial in healthcare, especially when leveraging data from multiple clinical sites to address data scarcity, reduce costs, and enable timely interventions. Transfer learning can facilitate cross-site knowledge transfer, but a major challenge is  heterogeneity in clinical concepts across different sites. 
Large Language Models (LLMs) show significant potential of capturing the semantic meaning of clinical concepts and reducing heterogeneity. 
This study analyzed electronic health records from two large healthcare systems to assess the impact of semantic embeddings from LLMs on local, shared, and transfer learning models. Results indicate that domain-specific LLMs, such as Med-BERT, consistently outperform in local and direct transfer scenarios, while generic models like OpenAI embeddings require fine-tuning for optimal performance. However, excessive tuning of models with biomedical embeddings may reduce effectiveness, emphasizing the need for balance. This study highlights the importance of domain-specific embeddings and careful model tuning for effective knowledge transfer in healthcare.}
%Knowledge transfer is vital in the healthcare domain, particularly when leveraging data from multiple clinical sites to address data scarcity or provide a timely response. Capturing semantic meaning from clinical notes is essential for knowledge transfer, where pre-trained Large Language Models (LLMs) can play a significant role. In this study, we used electronic health records from two large healthcare systems to test the impact of using semantic embeddings from LLMs on the performance of local models, shared models, and transfer learning models. Our results show that embeddings from LLMs pre-trained on biomedical knowledge, such as BioBERT and Med-BERT, consistently perform best in local and direct transfer learning scenarios. Meanwhile, representations from generic models like OpenAI embedding models require fine-tuning to achieve optimal performance. Interestingly, we also observe that extensive tuning of CNNs involving biomedical knowledge based embeddings can hurt their performance, highlighting the need for a balanced approach in transfer learning. This study underscores the importance of using domain-specific embeddings and careful model tuning to achieve successful knowledge transfer in healthcare.

\end{abstract} \hspace{10pt}


\section{Introduction}

\fontsize{10}{10}\selectfont
 
%In addition to the varied clinical documentation problems, data scarcity and privacy are also key concerns in the healthcare domain\cite{Rajendran2024}, regarding that,  transfer learning is a particularly valuable method to address the concerns \cite{Sun2024}. 
Effective knowledge sharing is vital in biomedicine, particularly when utilizing data from multiple clinical sites to overcome data scarcity, reduce computational costs, and ensure timely interventions. Transfer learning, an extended concept of traditional machine learning, can facilitate knowledge from one domain (e.g., one hospital) to be applied to tasks in a related but different domain (e.g., another hospital) \cite{pan2010survey,weiss2016survey, bao2024recentsurveyheterogeneoustransfer}. Collecting sufficient data on specific diseases for machine learning tasks in one site can be costly and time-consuming. Other clinical sites that have relevant disease cases can serve as the source sites to share the knowledge to help the target site build the models by the transfer learning algorithm \cite{Ji2023}. This approach reduces the cost and time, acquires the knowledge shortly to realize a timely response. Knowledge from different sites can be shared either by exchanging data or by sharing pre-trained models. The latter approach is particularly suitable for healthcare, as sharing pre-trained models helps maintain data confidentiality, addressing data privacy concerns \cite{Rajendran2024} across sites.

One challenge of effective transfer learning is to handle heterogeneity across domains \cite{bao2024recentsurveyheterogeneoustransfer}. In the biomedical domain, clinical concepts like symptoms or diseases are often expressed by diverse clinical languages \cite{Sohn2017}, including but not limited to free text, keywords, and different coding labels. For example, “fever” can be referred to as “high temperature” or labeled as “C0015867” in the UMLS CUI system \cite{Bodenreider2004-fq}. %This variability in clinical documentation introduces challenges and similar concepts may be ignored because of different coding systems. 
Relying on exact word matches to retrieve relevant clinical cases may not be sufficient. Such challenges become more pronounced in collaborative studies across institutions, where clinical practices often differ \cite{Sohn2017}. Therefore, to avoid missing the related features, learning the word representation is vital in clinical applications.

With the advancement of natural language processing (NLP), word embeddings, representing words as vectors to enable a measure of similarity, have become widely adopted methods for capturing semantic meaning \cite{Chandra2021}. Several pre-trained large language models (LLMs) have been developed to generate word embeddings, which have proven to be a promising method for NLP tasks in recent years. One notable LLM is Bidirectional Encoder Representations from Transformers (BERT) \cite{devlin2019}, which has become a popular model to learn linguistic knowledge. In the biomedical domain, BERT has been further adapted into specialized models, including 
%ClinicalBERT \cite{huang2019}, 
BioBERT \cite{lee2019}, and Med-BERT \cite{rasmy2021}. Meanwhile, OpenAI, the company that released ChatGPT, also introduced their third-generation text embedding models for tasks such as search, classification, and similarity measurement \cite{openai}. 

\iffalse
There are also existing studies that have discussed the application of NLP methods and LLMs with transfer learning in general areas \cite{lin2020exploring, chronopoulou2019embarrassingly,alyafeai2020surveytransferlearningnatural}, which focus on enhancing specific LLM performance or summarizing the algorithms. Its application to the biomedicine domain has also been studied, including topics of Electronic Health Records (EHRs) \cite{laparra2021review}, ECG diagnosis \cite{qiu2023transfer}, verbal autopsy report 
 \cite{Manaka2024}, and multimodal dataset \cite{belyaeva2024}. These studies mainly discussed the application of general NLP methods but failed to conclude the ability of different pre-trained LLMs, particularly overlooking the emergence of more clinically-focused LLMs.
 \fi

Previous research has explored the application of NLP methods and LLMs with transfer learning across various general tasks, often focusing on improving the performance of specific LLMs or summarizing related algorithms \cite{lin2020exploring, chronopoulou2019embarrassingly, alyafeai2020surveytransferlearningnatural}. In the biomedical field, similar approaches have been applied to areas such as Electronic Health Records (EHRs) \cite{laparra2021review}, ECG diagnosis \cite{qiu2023transfer}, verbal autopsy reports \cite{Manaka2024}, and multimodal datasets \cite{belyaeva2024}. However, these studies primarily address the general application of NLP techniques and do not comprehensively evaluate the capabilities of various pre-trained LLMs, particularly overlooking the emergence of more clinically-focused LLMs. Our work specifically fills this gap by providing a detailed analysis of how these advanced, clinically-tuned LLMs perform in biomedical applications, thus offering new insights that are not covered in existing literature.

In this study, we evaluated the benefits of using pre-trained LLM embeddings for local modeling and model-based transfer learning across institutions. Our investigation focused on two key questions:
\begin{itemize}
    \item Can semantic embeddings of clinical concepts improve the performance of classification tasks?
    \item How does transfer learning perform when using different types of semantic embeddings?
\end{itemize}

%We conducted experiments to develop models that detected influenza cases from EHR data from two geographically distinct institutions, demonstrating the potential of semantic representations to enhance classification tasks. Additionally, we assessed the importance of tuning strategies for appropriate LLMs to fully leverage their benefits in different scenarios.

We developed models to detect influenza cases from EHR data from two geographically distinct institutions, demonstrating the potential of LLM-based semantic representations to improve classification performance. Furthermore, we assessed the impact of various tuning strategies for optimizing LLMs, focusing on their effectiveness across different transfer learning scenarios and highlighting the importance of selecting the appropriate strategy to maximize their benefits.

%Semantic similarity is a key feature in many tasks related to texts, such as information retrieval, document classification, and summarization\cite{Kenter2015}. Traditional methods like Term Frequency Inverse Document Frequency
%\cite{Ramos2003} and Bag-of-Words \cite{zhang2010} were initially used to determine sentence or document relatedness by analyzing word frequency in texts. However, these approaches struggle to capture the semantic relationships between different words that may have similar meanings \cite{Chandra2021}. 

\section{Method}

\subsection{Overview}

\begin{figure}[!ht]
    \centering
    \includegraphics[width=0.7\linewidth]{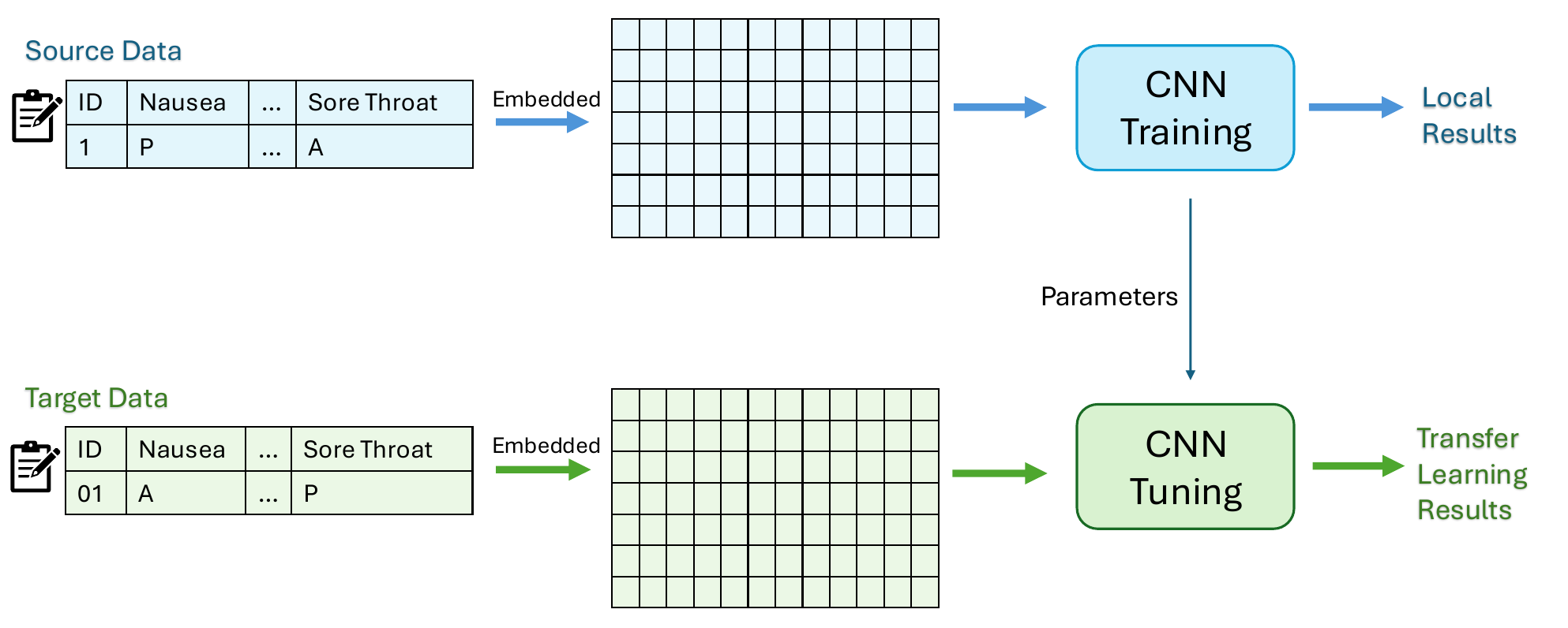}
    \caption{Workflow overview}
    \label{fig:overview}
\end{figure}

Figure \ref{fig:overview} shows the overall workflow in this study. 
%We transform tubular data into a 2D format, utilizing either One-Hot embeddings as a baseline or semantic embeddings. The embedded data serves as input for the convolutional neural networks (CNN) model, which is trained on this representation. For transfer learning experiments, the model is initially trained on the source dataset, and its parameters are later tuned on the target dataset. 
We project clinical concepts into embeddings (converting tabular data into a 2D format) using either a one-hot encoding method or embeddings from LLM models. The embedded data is then fed into a convolutional neural network (CNN) model, which is followed by a linear layer for classification tasks. CNNs were initially developed for image-related applications \cite{oshea2015introductionconvolutionalneuralnetworks}, and have been increasingly utilized in various NLP tasks \cite{minaee2021} since their first application to text classification tasks in 2014 \cite{Kim2014,kalchbrenner2014}. We conducted experiments on local or transfer learning scenarios. For local training, only the training data is used to learn the parameters, which are initialized randomly. In transfer learning experiments, the model is initially trained on the source dataset (e.g., the hospital providing knowledge) and then fine-tuned using the target dataset (e.g., the hospital receiving the knowledge). More details will be illustrated in the following experiment setting section (Section \ref{exp}). All experiment codes are available in \href{https://github.com/AI-In-Population-Health-Lab/CNN_package/tree/main}{this GitHub Repository}. \par

\begin{figure}[!ht]
    \centering    \includegraphics[width=0.9\linewidth]{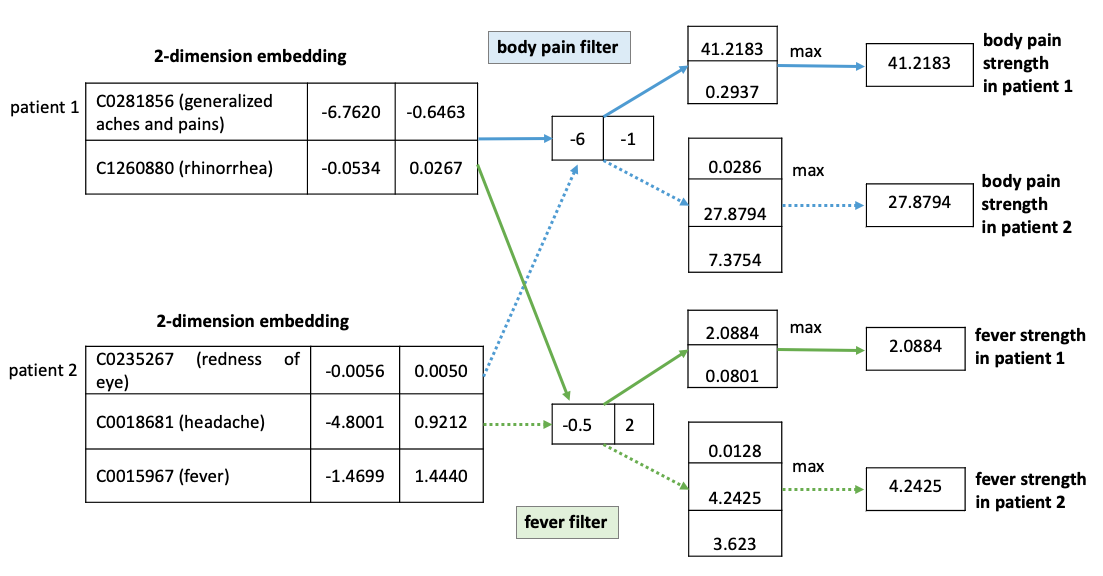}
    \vspace{-1 em}
    \caption{How embeddings, filters, and the max operation summarize patient information: Patient 1 presents two symptoms, and patient 2 presents three. Each symptom is represented by a vector with a length of 2. There are two tasks to detect ``body pain'' and ``fever'' respectively. The convolutional layer uses two corresponding filters with the same length of word representative vectors, and captures information related to each task. Blue arrows indicate calculations performed using the ``body pain filter.'' For Patient 1, the ``body pain filter'' outputs two values: (-6.7620) × (-6) + (-0.6463) × (-1) = 41.2183, and (-0.0534) × (-6) + (0.0267) × (-1) = 0.2937. For Patient 2, the ``body pain filter'' outputs three values: 0.0286, 27.8794, and 7.3754. The max pooling layer then selects the highest value, assigning 41.2183 as the body pain strength for Patient 1 and 27.8794 for Patient 2.}
 \label{fig:embedding}
\end{figure}

 Figure \ref{fig:embedding} illustrates two examples demonstrating how embeddings and CNN filters (within the convolution layer) can be used to summarize information in EHR notes. For the task of detecting ``body pain", the embedded symptoms in each patient's record are convoluted by the ``body pain filter," which detects pain in both patients but assigns a higher value to the pain in Patient 1's record. Similarly, a ``fever filter" identifies fever in Patient 2's record and assigns a higher value to fever strength. Even though fever is not explicitly mentioned in Patient 1's record, the presented symptom, generalized aches and pains, often co-occurs with fever in medical records, which mainly contributes to Patient 2's fever strength measure. This highlights the importance of capturing the semantic meaning of clinical concepts in EHR notes to help recognize synonyms and clinically related concepts. This approach also offers a significant advantage: when a model trained on source data is transferred to a target setting, these trained filters can still be used on the target variables, even if some variables do not appear in the source data. In this study, we use a 1-D convolutional layer to scan the clinical concept embeddings vertically, producing a vector of numerical outputs for the filter of the task on influenza cases classification.

% \subsection{CNN structure}
% This study applied CNN to the semantic meaning of the clinical concepts, which is a 1-D CNN with a single convolutional layer. The algorithm is illustrated in Figure \ref{fig:algorithm}, with an example of 7 clinical concepts and 5 embedding size. 
% Each tabular encounter data will be transformed into 2D embedded arrays with the shape of (feature size, embedded size) as our CNN model input. We applied 3 filters with sizes of 2, 3, and 4, number of 100 for each, padding of 0, stride of 1 on the convolutional layer, ReLU as activation function, 1 max-pooling layer, and a fully connected layer with a dropout of 0.5. The results will return binary to decide the classification.\par

\subsection{Data}
%https://journals.plos.org/plosone/article?id=10.1371/journal.pone.0174970
This study utilized Emergency Department encounter data with influenza diagnoses from Allegheny County (AC), Pennsylvania, and Salt Lake County (SLC), Utah, spanning 7 years from June 2008 to May 2015. The data distributions are detailed in Tables \ref{t1AC} and \ref{t1SLC}. The study is approved by the IRB of the
University of Pittsburgh (STUDY19050197). Positive cases refer to encounters confirmed by positive laboratory results, while negative cases are those confirmed by negative laboratory results.
The features used in this study include age group category (0-5, 6-64, 65+) and 70 clinical findings (Table \ref{t1FL}). We obtained these clinical concepts' values by applying an NLP tool, Topaz \cite{chapman2003creating}, to de-identified clinical notes from EHR. 
For each patient encounter, most clinical concepts were labeled as either ``present" or ``absent." The ``absent" label was applied when Topaz detected that the clinician explicitly documented the absence of a finding (e.g., ``patient denies cough") or when the finding was not mentioned in the clinical notes. The ``present" label was applied when Topaz detected that the clinician explicitly documented the presence of a finding. The highest measured temperature concept is with 4 potential labels: high grade ($>=$ 104.0F / 40C), low grade (100.4F–103.9F / 38–39.9C), inconsequential ($<$100.4F / 38C), and no temperature information. \par

%This study implemented Emergency Department encounter data with Influenza diagnosis from Allegheny County (AC), Pennsylvania and Salt Lake County (SLC), Utah, covering 7 years from Jun 2008 to May 2015. The data distributions are presented in Table \ref{t1AC} and \ref{t1SLC}. Positive cases are encounters that have been confirmed with positive laboratory results, while negative cases are encounters that have been confirmed with negative laboratory results. \par
%Among these concepts, the highest recorded temperature was categorized into three possible levels: high-grade ($\geq 104.0^\circ\text{F} / 40^\circ\text{C}$), low-grade ($100.4^\circ\text{F}–103.9^\circ\text{F} / 38–39.9^\circ\text{C}$), and inconsequential ($<100.4^\circ\text{F} / 38^\circ\text{C}$). 

The embeddings of clinical concepts involved in this study were extracted from two groups of LLMs (Table \ref{t1LLM}): BERT and OpenAI. The BERT group includes original BERT \cite{devlin2019}, trained from generic corpus, and its biomedical variants, such as BioBERT \cite{lee2019}, trained from both generic cohort and biomedical articles, and Med-BERT \cite{rasmy2021}, trained from EHR database and claims dataset. Though OpenAI did not public the training rationale or dataset for their latest third-generation embedding models, text-embedding-3-small and text-embedding-3-large \cite{openai} (referred to in this study as OpenAI-S and OpenAI-L, respectively), it can be inferred that they were trained on more generic datasets. For clinical concepts labeled as ``present," embeddings were extracted directly based on their semantic representation, while for those labeled as ``absent," embeddings were extracted based on their negation. Similarly, embeddings for age groups and temperature grade were extracted according to their respective semantic meanings. All embeddings from BERT were extracted by the [CLS] tokens, the aggregated representation of the entire input. OpenAI embedding models return straightforward sentence-level embeddings.

%The involved embeddings of clinical concepts are extracted from two LLM groups (Table \ref{t1LLM}): BERT and OpenAI. From BERT, there are extended LLMs trained in biomedical cohorts, such as BioBERT and Med-BERT. OpenAI's latest third-generation embedding models, text-embedding-3-small and text-embedding-3-large (referred to as OpenAI-S and OpenAI-L in this study, respectively), are trained on more generic datasets. The embeddings of Presented clinical concepts are extracted based on themselves, Absent clinical concepts are extracted based on their negation. The embeddings of age groups are extracted based on their semantic meaning respectively.  

\begin{table}[!htb]
    \begin{subtable}{.5\linewidth}
      \centering
        \begin{tabular}{|c|c|c|c|}
           \hline
    Admit Date&Total&Negative&Positive\\
    \hline
    20080601-20090531&338&293&45\\
20090601-20100531&3030&2338&692\\
20100601-20110531&1849&1522&327\\
20110601-20120531&1827&1784&43\\
20120601-20130531&2766&2369&397\\
20130601-20140531&2214&1924&290\\
20140601-20150531&3418&2965&453\\
    \hline
        \end{tabular}
        \caption{AC Data Distribution}
        \label{t1AC}
    \end{subtable}%
    \hspace*{\fill}
    \begin{subtable}{.5\linewidth}
      \centering
        \begin{tabular}{|c|c|c|c|}
        \hline
    Admit Date&Total&Negative&Positive\\
    \hline
    20080601-20090531&5500&5094&406\\
20090601-20100531&8649&7587&1062\\
20100601-20110531&6350&5693&657\\
20110601-20120531&5716&5375&341\\
20120601-20130531&8064&7040&1024\\
20130601-20140531&6848&6423&425\\
20140601-20150531&8437&7798&639\\
    \hline
        \end{tabular}
        \caption{SLC Data Distribution}
        \label{t1SLC}
    \end{subtable} 

\begin{subtable}{.5\linewidth}
      \centering
        \begin{tabular}{|c|c|}
        \hline
    Features&Labels\\
    \hline
    \multirow{2}{*}{69 Clinical Concepts}& P (Present)\\ 
    &A (Absent)\\ \hline
    \multirow{2}{*}{Temperature}& High grade, Low grade, \\
    &Inconsequential, No info \\ \hline
    \multirow{4}{*}{Age Group}& 0-5\\
    &6-64\\
    &65+\\
    \hline
        \end{tabular}
        \caption{Features and Labels}
        \label{t1FL}
    \end{subtable} %
    \hspace*{\fill}
\begin{subtable}{.5\linewidth}
      \centering
        \begin{tabular}{|c|c|c|}
        \hline
    Group&LLMs& Knowledge\\
    \hline
    \multirow{5}{*}{BERT} & \small BERT & \small Books Corpus, English Wikipedia\\ \cline{2-3}
    &\multirow{2}{*}{\small BioBERT}&  \small English Wikipedia,
PubMed Abstracts\\ 
&& \small Books Corpus, PMC Full-text articles\\ \cline{2-3} 
    &\multirow{2}{*}{\small Med-BERT}&\small Cerner Health Fact\\
    &&\small Truven Health MarketScan\\  \hline
    % \multirow{2}{*}{Llama}& Llama\\
    % &Me-Llama\\ \hline
    \multirow{2}{*}{OpenAI}& \small OpenAI-S&\multirow{2}{*}{\small Not disclosed} \\
    &\small OpenAI-L &\\
    \hline
        \end{tabular}
        \caption{LLMs and Corresponding Knowledge}
        \label{t1LLM}
    \end{subtable} 
    
    \caption{Data Description}
\end{table}

\newpage
\subsection{Experiment Settings}
\label{exp}

To verify the ability of pre-trained semantic embeddings, we conduct experiments with One-Hot encoding as our baseline. Only pointing out the features' presence, One-Hot encoding has no semantic meaning at all. To further compare the performance of embeddings from different LLMs, we involve both generic and biomedical-extended LLMs from two algorithm based groups - BERT and OpenAI embeddings. In order to explore the transferability of pre-trained embeddings, each embedding is applied to train the local and shared model under different transfer learning structures.

\begin{figure}[!ht]
    \centering
    \includegraphics[width=0.9\linewidth]{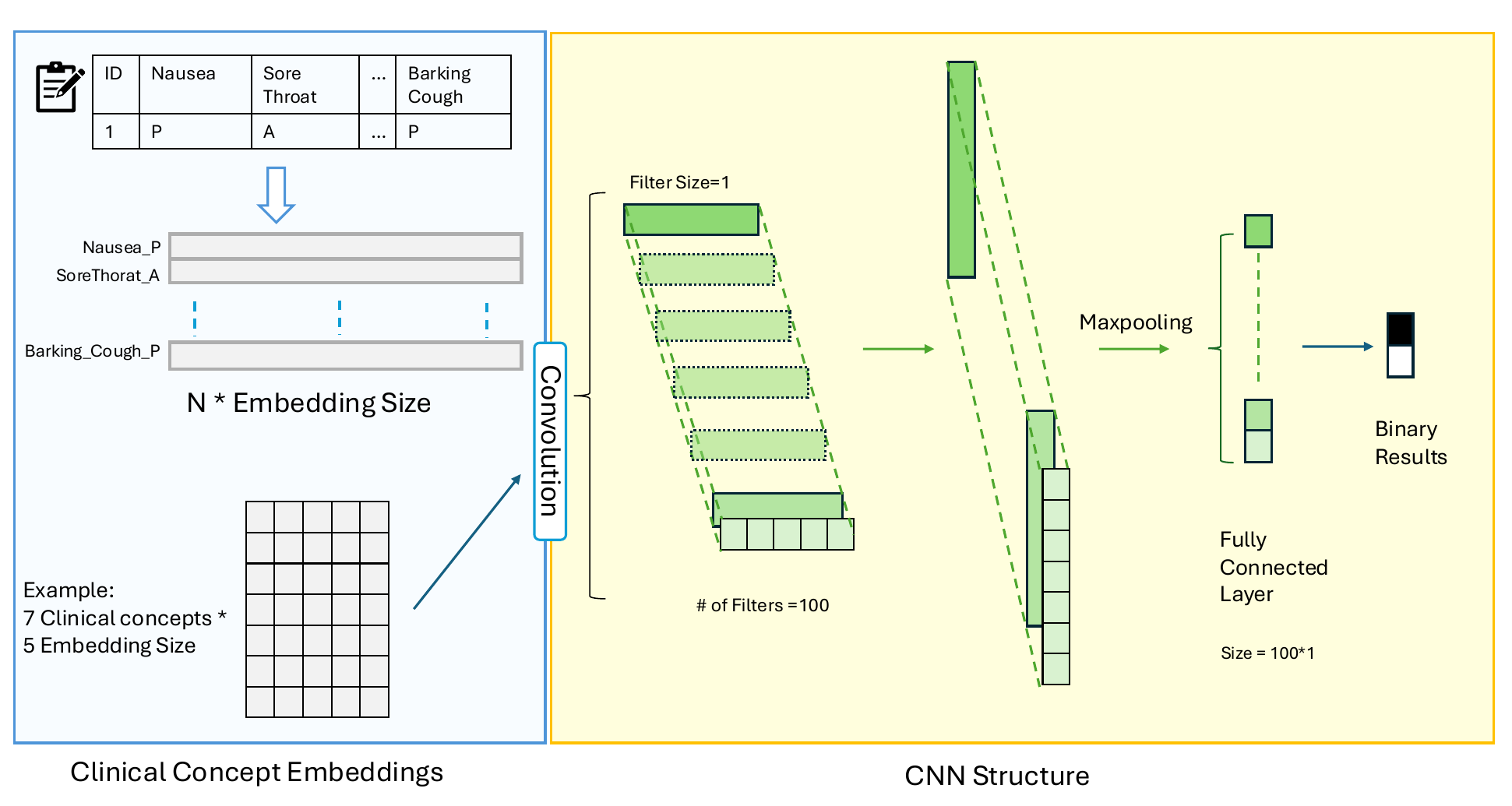}
    \caption{Clinical Concepts Embedding CNN Struture}
    \label{fig:algorithm}
\end{figure}

All tabular data are first converted into 2D embeddings along with their labels using either One-Hot encoding or pre-trained semantic embeddings from LLMs. This process transforms the input data to a size of (Number of clinical concepts × Embedding Size). In this study, a single 1-D convolutional layer is applied, using a stride of 1, no padding, and the filter size of 1 with 100 filters. The size of filters is depicted on a height equal to 1, where the width is equal to the size of embeddings, we refer to the height for the filter size \cite{zhang2017sensitivity}. Setting the filter size as 1 allows the filter to distinguish each row respectively, avoiding the influence of the order of clinical concepts. The ReLu activation function is used, followed by a maxpooling layer and a fully connected layer. Dropout is set to 0.5 to prevent overfitting. The learning rate is set to 0.001 to ensure effective convergence without overshooting. Figure \ref{fig:algorithm} provides a detailed overview of the experiment structure, illustrated with an example of 7 clinical concepts and 5 embedding sizes. Table \ref{tab:parametersl}  presents the embedding sizes and the number of parameters in each layer of corresponding models. To compare the different pre-trained embeddings intuitively, we freeze the embedding layer to fix the original embeddings.

%All tabular data are first embedded into 2D dimensions with their labels by One-Hot embedding or semantic embeddings pre-trained from LLMs, so that the size of input data becomes (Number of clinical concepts $\times$ Embedding Size). This study applied a single 1-D convolutional layer, with the stride of 1, padding of 0, and 3 sizes of filters (2,3,4 respectively), and each size has a number of 100 filters. Rectified Linear Unit (ReLU) is adapted as the activation function, followed by one maxpooling layer and one fully connected layer. Dropout is set as 0.5. Learning rate is set as 0.001, allowing the model to converge effectively without the risk of overshooting the optimal solution. Figure \ref{fig:algorithm} demonstrates the detailed experiment structure, with a simple example of 7 clinical concepts and 5 embedding sizes. Table \ref{tab:parametersl} presents the embedding sizes and the number of parameters in each layer of corresponding models. To compare the different pre-trained embeddings intuitively, we freeze the embedding layer to fix the original embeddings.

%Comparing the transferability of models between sites, this study conducts bi-direction experiments, namely, AC and SLC have been selected as target sites respectively, which adopts the knowledge from the other site.Given the time-sensitive characteristic of encounter data, our training, validation, and test data are split based on the admitted date when developing models.The models have been developed in the following two scenarios:

This study compares the transferability of models between two sites by conducting bi-directional experiments. Either AC or SLC has been selected as the target sites, with each site adopting knowledge from the other. Due to the time-sensitive nature of encounter data, we split the training, validation, and test datasets based on the admission date during model development. The models are developed in two scenarios:
\begin{itemize}
    \item Local (AC-only or SLC-only) CNN classification model ;
    \item Transfer learning CNN classification model.
\end{itemize}

% Hypothesis:
% do not emphasize too much about CNN because now transformer is dominate \\
% 1. llm-embedding is useful. llm $>$ one hot. local comparison. 
% AC: llm + CNN $>$ one hot + CNN, BN\\
% 2. llm + TL $\ge$ llm + local (few target data) $\ge$ one hot + TL \\

% hyper, train, val, test (5)

% Local-CNN $>$  BN, CNN can get semantic meaning across concepts.
% local-CNN has limited data, TL better; local-CNN has enough data, TL =local.
% MLP, wo embedding
% TL-CNN  >=local-CNN, TL: limited local data, 

In each scenario, data from the target site is divided into a training-validation set (June 2008–May 2014) and a test set (June 2014–May 2015). The training-validation set is further randomly split in an 8:2 ratio to define the training and validation datasets. In the transfer learning scenario, data from the source site is included only for training and validation purposes, covering the same period (2008–2014), and is also split randomly in an 8:2 ratio. The model parameters trained on the source site data are loaded before tuning on the target site data. Data allocations for the different models are detailed in Table \ref{modelData}. \par

\begin{table}[]
    \centering
    \begin{tabular}{|c|c|c|c|c|}
    \hline
    \multirow{2}{*}{Model}& \multirow{2}{*}{Embedding Size }&\multicolumn{2}{|c|}{Number of Parameters}\\ \cline{3-4}
    & & Convolutional Layer & Fully Connected Layer\\ \hline
    One-Hot & 145 & 14,500& 202\\
    BERT-based & 768&76,800& 202  \\
    OpenAI-S  & 1536&153,600 & 202\\
    OpenAI-L & 3072&307,200 &202 \\
    \hline
    \end{tabular}
    \caption{Embedding Size and Number of Parameters}
    \label{tab:parametersl}
\end{table}

% \begin{table}[]
%     \centering
%     \begin{tabular}{|c|c|c|c|c|}
%     \hline
%     \multirow{2}{*}{Model}& \multirow{2}{*}{Embedding Size }&\multicolumn{3}{|c|}{Number of Parameters}\\ \cline{3-5}
%     & &Embedding Layer (Freeze) & Convolutional Layer & Fully Connected Layer\\ \hline
%     One-Hot & 145& 10,295 & 14,500& 202\\
%     BERT-based & 768& 54,528&76,800& 202  \\
%     OpenAI-S  & 1536& 109,056&153,600 & 202\\
%     OpenAI-L & 3072 & 218,112&307,200 &202 \\
%     \hline
%     \end{tabular}
%     \caption{Embedding Size and Number of Parameters}
%     \label{tab:parametersl}
% \end{table}

\begin{table}[h!]
  \centering
  \begin{tabular}{|c|c|c|c|}
    \hline
    Scenarios&Local Training and Validation&Local Testing& Source Training and Validation\\
    \hline
    Local-CNN& AC (or SLC) 2008-2014& AC (or SLC) 2014-2015&/ \\
    TL-CNN& AC (or SLC) 2008-2014& AC (or SLC) 2014-2015&SLC (or AC) 2008-2014\\

    \hline
  \end{tabular}
  \caption{Data Allocations}
  \label{modelData}
\end{table}

\newpage
\section{Results}
% Expected results (all for transfer learning scenarios):
% \begin{itemize}
%     \item CNN-structure perform better than the baseline
%     % \item CUI description embedding performs better than CUI word embedding
%     \item CNN TL can reach a similar or higher performance than local CNN.
% \end{itemize}

\begin{figure}[!ht]
    \centering
\includegraphics[width=0.8\textwidth]{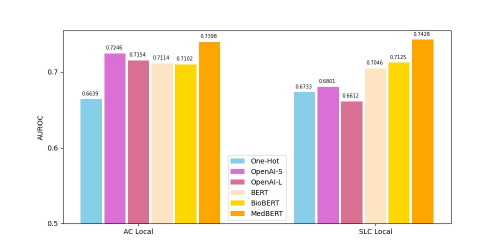}
    \caption{AUROCs of Local models on AC and SLC}
    \label{fig:1}
\end{figure}

Figure \ref{fig:1} presents the Area Under the Receiver Operating Characteristic curves (AUROC) values for models developed only in the local scenario. As shown by the bar chart, the BERT-based models consistently achieved high AUROCs in both local settings. The embeddings from Med-BERT, pre-trained with clinical knowledge, yielded the highest AUROCs compared to all other embeddings, including the biomedical knowledge based BioBERT. In the AC Local setting, the embeddings from OpenAI models outperformed the One-Hot model, which lacks semantic meaning, while presented an AUROC close to the One-Hot embeddings in the SLC Local setting.

%Table \ref{tab:res1} and Table \ref{tab:res2} show Area Under the Receiver Operating Characteristic curves (AUROCs) of models developed under different settings. In AC Local, One-Hot, as the baseline model with no semantic meaning, presented an AUROC  close to the OpenAI groups, while in SLC Local, One-Hot delivered a higher performance than OpenAI models. The BERT-based group consistently scored the highest in both local situations, where the generic semantic embedding of BERT presented lower AUROCs than the embeddings pre-trained with biomedical knowledge, BioBERT and Med-BERT. This trend also can be seen from the underlined in the first columns (Local Models) of Table \ref{tab:res1} and Table \ref{tab:res2}. 

Tables \ref{tab:res1} and \ref{tab:res2}, present the AUROC performance for models developed under different settings, where the first column depicts the same local results as Figure \ref{fig:1} and the later columns depict the results in sharing the source models.  To compare the transferability of source-trained models to the target site, we applied three adaptation methodologies: direct sharing, tuning only the linear layers, and tuning both convolutional and linear layers. As shown in the second columns of Tables \ref{tab:res1} and \ref{tab:res2}, when directly sharing source-trained models with a new healthcare system, models using semantic embeddings consistently outperformed the baseline models. Among them, it was also Med-BERT to achieve the highest AUROC values like in local scenarios, marked by the underline in the first two columns.

%When comparing the transferability of source-trained models on the target site, we adapted source-trained models with three methodologies - directly shared, tuned on linear layers, and tuned both convolutional and linear layers. From the second columns of Table \ref{tab:res1} and Table \ref{tab:res2}, it is evident that when directly sharing the source-trained models with a new healthcare system, the models adopted biomedical embeddings still score higher than baseline or those with generic embeddings, as the underline marks the highest AUROC within this column. 

%When tuned on linear layers, the most shared models get their performance improved. Tuning on both convolutional and linear layers has further enhanced the performance of baseline and OpenAI models, but surprisingly, harms the results in BERT groups. The bolded AUROCs represent the highest scores from the same embedding settings compare across the columns in each row. Baseline and OpenAI models have their best performance after tuned on both convolutional and linear layers, while BERT-based models prefer to have their best performance more in tuning on linear layer only and, in some cases, in the directly shared situation.

When tuning only the linear layers, most shared models showed improved performance. Further tuning both convolutional and linear layers enhanced the performance of the baseline (one-hot encoding) and OpenAI models. However, unexpectedly, this approach negatively impacted the performance of the BERT-based models. The bolded AUROCs indicate the highest scores within the same embedding settings across the columns in each row under transfer learning scenario. Baseline and OpenAI models achieved their best performance when both convolutional and linear layers were tuned, while BERT-based models generally performed best when only the linear layers were tuned, or in some cases, when the models were directly shared. The full tuning brought limited enhancement or even hurt the performance.

Overall, our experiment results demonstrate that embedding models with clinical knowledge, such as Med-BERT, significantly enhances both local performance and the portability of models to new sites, often preserving high performance with minimal tuning. Conversely, when domain-specific pre-trained models are not available, transfer learning with one-hot encoding or general embeddings can achieve comparable performance, but this typically requires more extensive model tuning with data from the target sites.

% \begin{table}[h!]
%   \centering
%   \begin{tabular}{|c|c|c|c|c|}
%     \hline
%     &Local Models (AC) & Shared Models (SLC) & Tune Linear layers & Tune CNN and linear layers\\
%     \hline
%     One-Hot & 0.7165& 0.7307 & 0.7330& \textbf{0.7421} \\
%     OpenAI-Small &  0.7156 &0.7147&0.7225& \textbf{0.7366} \\
%     OpenAI-Large&  0.7152 &0.7019&0.7082& \textbf{0.7442}\\
%     BERT & 0.7277&0.7363& \textbf{0.7400}& 0.6749\\
%     BioBERT&  \underline{0.7316} &\textbf{0.7475}&0.7425& 0.7240\\
%     Med-BERT&  \underline{0.7316}&\underline{\textbf{0.7501}}&0.7351& 0.7234 \\ 
%     \hline
%   \end{tabular}
%   \caption{Performance of models when transferring from SLC to AC}
%   \label{tab:res1}
% \end{table}

% \begin{table}[h!]
%   \centering
%   \begin{tabular}{|c|c|c|c|c|}
%     \hline
%     &Local models (SLC) & Shared models (AC) & Tune linear layers & Tune CNN and linear layers\\
%     \hline
%     One-Hot & 0.7311&0.7196&0.7403& \textbf{0.7518} \\
%     OpenAI-Small &  0.7142 &0.7161&0.7339& \textbf{0.7512} \\
%     OpenAI-Large&  0.7025 &0.7141& 0.7280& \textbf{0.7526}\\
%     BERT & 0.7355&0.7281& \textbf{0.7428}& 0.7121\\
%     BioBERT&  0.7470 &\underline{0.7348}& \textbf{0.7474}& 0.7363\\
%     Med-BERT&  \underline{0.7491}&0.7334&\textbf{0.7494}&0.7431 \\ 
%     \hline
%   \end{tabular}
%   \caption{Performance of models when transferring from AC to SLC}
%   \label{tab:res2}
% \end{table}

\begin{table}[h!]
  \centering
  \begin{tabular}{|c|c|c|c|c|}
    \hline
    &Local Models (AC) & Shared Models (SLC) & Tune Linear layers & Tune CNN and linear layers\\
    \hline
    One-Hot&0.6639&0.6886&	0.6924	&\textbf{0.7322}\\
OpenAI-S&	0.7246&	0.7083&	0.7117&	\textbf{0.7351}\\
OpenAI-L&	0.7154&	0.6928&	0.6984&	\textbf{0.7360}\\
BERT&	0.7114&	0.7031&	\textbf{0.7363}&	0.6843\\
BioBERT	&0.7102&\textbf{0.7189}&	0.6840&	0.7167\\
Med-BERT&	\underline{0.7398}&	\underline{0.7308}&	\textbf{0.7319}&	0.7315\\
    \hline
  \end{tabular}
  \caption{Performance of models when transferring from SLC to AC}
  \label{tab:res1}
\end{table}

\begin{table}[h!]
  \centering
  \begin{tabular}{|c|c|c|c|c|}
    \hline
    &Local models (SLC) & Shared models (AC) & Tune linear layers & Tune CNN and linear layers\\
    \hline
    One-Hot&0.6733	&0.6655&	0.7057&	\textbf{0.7515}\\
OpenAI-S&0.6801	&0.7077&	0.7086&	\textbf{0.7475}\\
OpenAI-L&0.6612	&0.7105	&0.7196&	\textbf{0.7498}\\
BERT&0.7046&	0.7032&	\textbf{0.7158}&	0.7030\\
BioBERT	&0.7125&	0.7053&	0.7002&	\textbf{0.7190}\\
Med-BERT&	\underline{0.7428}&	\underline{0.7436}&	\textbf{0.7477}&	0.7314\\
    \hline
  \end{tabular}
  \caption{Performance of models when transferring from AC to SLC}
  \label{tab:res2}
\end{table}

\section{Discussion}

There is a clear trend from our results: pre-trained embeddings with clinical knowledge, like Med-BERT, consistently outperformed other models, in local scenario or even when directly adopting models trained from a different healthcare system. This highlights that the medical knowledge augmented embedding performs better in local modeling than generic embeddings and also are more robust when sharing to a new healthcare system. And the generic model BERT also exhibits higher AUROC than the baseline, which indicates semantic embeddings can be beneficial for classification tasks on EHR data. We also compared the performance of another biomedical knowledge involved LLM, BioBERT, which exhibited the results much lower than Med-BERT. The different training corpus may be the contributor to this situation. BioBERT is trained on both generic corpus and biomedical research articles, while Med-BERT is trained on the EHR database and the medical claims dataset. Therefore, Med-BERT has a more clinical-specific context than BioBERT, enabling it to adapt well to this EHR-based study. 
% As BERT is trained bidirectionally using masked language model algorithm \cite{devlin2019},  the word representations from BERT can become more context-specific than other language models \cite{ethayarajh2019contextual}. Even though OpenAI did not disclose the rationale for how they trained their embedding models, it can be inferred that they have less medical context knowledge than BERT does. \par

However, when we start to tune the shared models, the advantages of clinical knowledge-based embeddings become ambiguous. Med-BERT still outperformed when tuning only happens on the linear layer, but full tuning on both convolutional and linear layers negatively impacts the performance, in contrast, the baseline and OpenAI models benefit from the full tuning. Here brings a question: after sharing, whether the model needs to be tuned, and to what extent? Considering the CNN structure, the convolutional layer focuses on capturing the semantic meaning of embedded EHR documents, and the fully connected linear layer may be more related to local tasks. For instance, the parameters of the linear layer can be distinct when this influenza diagnosis is sampled on a random group of people or influenza-symptomed patients. This suggests that domain-specific models may already have robust representations in their embeddings, requiring minimal fine-tuning to adapt to new domains. In contrast, models with less specialized knowledge, like the baseline or OpenAI models, require full tuning to achieve optimal performance. 
%Therefore, it can be implied that domain-specific models, like BioBERT and Med-BERT mayalready have robust representations in their embeddings, requiring less extensive fine-tuning to adapt to new domains; while if the knowledge is not sufficient, like the baseline or OpenAI models, the full tune is desired to reach the ideal performance. 
In addition, it has been demonstrated that BERT is better adapted in a lightweight transfer learning model \cite{peters2019tune}, so that full tuning may hurt the performance of BERT-based embeddings.
% A study has pointed out, that the word representations from BERT can become more context-specific, but GPT produces representations that are no more similar than those from randomly sampled words \cite{ethayarajh2019contextualcontextualizedwordrepresentations}.
 % A previous study also demonstrated that BERT is better adapted in a lightweight transfer learning model, the increasing parameter hurts the performance\cite{peters2019tunetuneadaptingpretrained}, which can be the possible reason to explain the decreased AUROC in our model as well.\par

While this study demonstrates the importance of using semantic embeddings for EHR classification, a few limitations exist. \textbf{First}, we employed simple One-Hot encodings as a baseline for binary classification tasks and acquired good performance after tuning. However, more complex clinical tasks, such as disease stage classification or multimodal data analysis, may not perform well with One-Hot encodings. These tasks may require more sophisticated representations to capture hidden patterns. \textbf{Second}, we only compared two groups of pre-trained LLMs, BERT-based models and OpenAI models, from which OpenAI has not disclosed their rationale. There are many other LLMs that have demonstrated success not only in generic tasks but also in the biomedical domain, like Llama \cite{touvron2023llamaopenefficientfoundation} (Me-Llama as its biomedical variant \cite{xie2024me}), 
%or GPT-based methods \cite{kalyan2023survey}, 
yet their transferability and sensitivity on tuning process remains untested. \textbf{Third}, the embeddings of absent clinical concepts were retrieved based on their negations, which may direct a less-accuracy semantic capture. Embeddings from detailed EHR, like free-text, may help to capture the more accurate representations.  \textbf{Fourth}, given the task in this study is relatively simple, the structure of the CNN model was arbitrarily designed and the hyperparameters were not extensively explored, for example, the embedding layer was frozen for pre-trained embeddings direct-apply. It’s possible that the model can achieve a higher performance with more dedicated settings.  \par

\begin{comment}
- limitation: \\
1. limitation, embedding not tune, future work, tune embedding, efficient loRA)
2. CNN structure design arbitrary, (simple task, data instances not too much), future transformer structure
\end{comment}

Our results have informed that the fine-tuning strategy should be tailored to the specific characteristics of the pre-trained embeddings used. Future research should further compare the semantic representation of different LLMs on clinical data and explore efficient model structure to improve the adaptability. Combining LLMs with lightweight model tuning methods, like adapter modules \cite{houlsby2019parameter} or low-rank adaptation (LoRA) \cite{hu2021loralowrankadaptationlarge}, could benefit the efficient adaptation to specific clinical systems with minimal computational cost.  Lastly, while LLMs hold great promise in the healthcare domain, the strategy to select LLMs may depend on factors such as dataset characteristics, local tasks or policies, and the underlying structure of the LLM.

\begin{comment}

\textcolor{red}{
RB comments:
\begin{itemize}
    \item It would be better to outline the sequence and rationale behind each experimental step, particularly regarding the fine-tuning process. Itemize them to be organized if possible
    \item Include detailed information on how hyperparameters were selected, such as filter sizes and learning rates. (YG: added one sentence about learning rate; other hyperparameters were select arbitrarily )
    \item When discussing, provide a brief introduction to the different LLM models to help readers better understand why the BERT-based model performs the best. (YG: added description about BERT)
    \item Consider using charts (e.g., bar graphs) to make comparisons more visually clear.
     \textbf{Current bar charts shows that the difference is trivial.see figure 4}
    \item In the discussion/conclusion, discuss how the findings can inform real-world model/applications and fine-tuning choices. (YG: added in and modified the last paragraph in Discussion)
\end{itemize}
}
\end{comment}

\section{Conclusion}
% 1. local: complex embeddings may be better
% 2. share purpose: simpler embeddings; large embeddings may lead to overfitting (not easy share)
% 3. different institution share, institution yearly updating, online learning, 
% domain generation (simple embedding, essential knowledge may be more effective, stable, model calculation easy).

% Simultaneously tuning only the connecting layers (while keeping embeddings fixed) does not allow the model to fully adapt to the new hospital's unique data distribution and context.

Semantic embeddings play a crucial role in the biomedical domain for both local tasks and the effectiveness of transfer learning. More sophisticated, context-rich embeddings from domain-specific models, like Med-BERT, generally provide better  performance, higher portability, and require less extensive tuning. Fine-tuning strategies—whether focusing on linear layers or combining CNN and linear adjustments—depend on the complexity and specialization of the model, as well as the similarity between source and target domains. Understanding these dynamics is key to optimizing knowledge sharing across different clinical sites.

% By comparing the performance of different LLMs in local scenarios, this study has demonstrated that involving semantic representations with EHR documents can improve the performance of classifying disease cases in the specific clinical site, and the embeddings pre-trained with clinical knowledge, like BioBERT and Med-BERT can further enhance the results, indicating the potential of pre-trained LLMs' application in the biomedical domain. However, in transfer learning scenarios, the BERT-based model delivered poor results, bringing a new question: with the contextualized structure, BERTs can benefit from their domain-specific richness, but may not have satisfactory generalizability. For transfer learning between different sites, simpler embeddings (like One-Hot or OpenAI-Small) are more effective, contributed by their generalized representations that adapt more readily to the new domain's specific characteristics. The selection of LLMs for future application in real-world biomedical domains can be one of the research directions.

\section{Acknowledgement}
This work was supported by the research grant
R00LM013383 from the National Library of Medicine, National Institutes of Health.

\titleformat*{\section}{\fontsize{10}{12}\bfseries\centering}
\bibliography{literature}

\end{document}